\documentclass[sigconf]{acmart}
\AtBeginDocument{%
  }

\setcopyright{acmlicensed}
\copyrightyear{2018}
\acmYear{2018}
\acmDOI{XXXXXXX.XXXXXXX}
\acmConference[Conference acronym 'XX]{Make sure to enter the correct
  conference title from your rights confirmation email}{June 03--05,
  2018}{Woodstock, NY}
\acmISBN{978-1-4503-XXXX-X/2018/06}

\usepackage{float}
\usepackage{multirow}
\usepackage{graphicx}
\usepackage{subcaption}

\begin{document}

\title{MoE-Health: A Mixture of Experts Framework for Robust Multimodal Healthcare Prediction}


\author{Xiaoyang Wang}
\email{xw388@drexel.edu}
\orcid{0000-0002-8471-4670}

\affiliation{%
  \institution{Drexel University}
  \city{Philadelphia}
  \state{PA}
  \country{USA}
}

\author{Christopher C. Yang}
\email{chris.yang@drexel.edu}
\orcid{0000-0001-5463-6926}
\affiliation{%
  \institution{Drexel University}
  \city{Philadelphia}
  \state{PA}
  \country{USA}
}

% \author{Anonymous Submission}
% \email{chris.yang@drexel.edu}
% \orcid{0000-0001-5463-6926}
% \affiliation{%
%   \institution{Drexel University}
%   \city{Philadelphia}
%   \state{PA}

\renewcommand{\shortauthors}{Wang et al.}

\begin{abstract}
Healthcare systems generate diverse multimodal data, including Electronic Health Records (EHR), clinical notes, and medical images. Effectively leveraging this data for clinical prediction is challenging, particularly as real-world samples often present with varied or incomplete modalities. Existing approaches typically require complete modality data or rely on manual selection strategies, limiting their applicability in real-world clinical settings where data availability varies across patients and institutions. To address these limitations, we propose MoE-Health, a novel Mixture of Experts framework designed for robust multimodal fusion in healthcare prediction. MoE-Health architecture is specifically developed to handle samples with differing modalities and improve performance on critical clinical tasks. By leveraging specialized expert networks and a dynamic gating mechanism, our approach dynamically selects and combines relevant experts based on available data modalities, enabling flexible adaptation to varying data availability scenarios. We evaluate MoE-Health on the MIMIC-IV dataset across three critical clinical prediction tasks: in-hospital mortality prediction, long length of stay, and hospital readmission prediction. Experimental results demonstrate that MoE-Health achieves superior performance compared to existing multimodal fusion methods while maintaining robustness across different modality availability patterns. The framework effectively integrates multimodal information, offering improved predictive performance and robustness in handling heterogeneous and incomplete healthcare data, making it particularly suitable for deployment in diverse healthcare environments with heterogeneous data availability.
\end{abstract}


% \ccsdesc[500]{Computing methodologies~Machine learning approaches}


\received{20 February 2007}
\received[revised]{12 March 2009}
\received[accepted]{5 June 2009}

\maketitle

\section{Introduction}
The integration of artificial intelligence in healthcare has witnessed remarkable progress, with predictive models increasingly leveraging diverse data modalities to improve clinical decision-making~\cite{soenksen2022integrated}. Modern healthcare systems generate vast amounts of heterogeneous data, including structured electronic health records (EHRs), unstructured clinical notes, and medical imaging data~\cite{soenksen2022integrated,shaik2024survey}. This multimodal nature of healthcare data presents both opportunities and challenges for developing robust predictive models that can support critical clinical tasks such as mortality prediction, length of stay estimation, and readmission forecasting.

While multimodal approaches have demonstrated superior performance compared to single-modality methods, existing frameworks face significant limitations in real-world deployment scenarios. Most current multimodal healthcare prediction models are trained and evaluated on patient subsets with complete modality availability~\cite{wang2024multimodal,wang2024multimodalfusion}, overlooking the variability and incompleteness that are inherent in clinical practice. In reality, healthcare institutions differ in their data collection protocols, technological infrastructure, and documentation quality, resulting in frequent cases where certain modalities are unavailable. Our analysis of the MIMIC-IV dataset illustrates this issue clearly: among 31,088 admissions, only 37.4\% contain all three modalities—structured EHR, clinical notes, and CXR images. This missingness highlights a key limitation of existing approaches: by assuming fully observed inputs, they exclude a large proportion of the patient population and fail to model the complex interactions between partially observed modalities.

Such limitations are particularly problematic in high-stakes environments like emergency departments, where timely decisions must be made based on whatever data is immediately accessible~\cite{zhang2022m3care,arnaud2023predictive}. In these scenarios, reliance on fully observed data is impractical. Therefore, there is a pressing need for predictive frameworks that are robust to missing or varying modalities, capable of adapting dynamically to whatever information is available. These models must ensure reliable decision support across the full spectrum of patients, rather than a privileged subset with complete records.

To address these challenges, we propose MoE-Health, a novel Mixture of Experts (MoE) framework specifically designed for robust multimodal healthcare prediction. MoE-Health leverages the strengths of expert models specialized for different data modalities (or combinations thereof) and a gating mechanism to dynamically weigh their contributions. Specifically, this architecture allows the framework to effectively handle samples with differing sets of available modalities and learn tailored fusion strategies. 

The main contributions of this work are as follows:
\begin{itemize}
    \item We propose MoE-Health, a Mixture of Experts framework specifically engineered for robust clinical prediction from heterogeneous multimodal data with inherent missingness.
    \item We design a unique architecture featuring specialized experts, each pre-trained to handle specific combinations of data modalities, coupled with a dynamic gating mechanism that intelligently routes patient data. This design explicitly models and adapts to real-world data availability patterns.
    \item We conduct extensive experiments on the large-scale MIMIC-IV dataset across three clinically significant prediction tasks. Our results validate that MoE-Health consistently outperforms established baselines, demonstrating the framework's effectiveness and robustness in realistic, incomplete-data scenarios.
\end{itemize}

\section{Related Work}
\subsection{Healthcare Predictive Modeling}
Predictive modeling in healthcare has seen significant advancements, particularly with the availability of large-scale datasets like MIMIC-IV~\cite{johnson2024mimic}. Traditional approaches, such as logistic regression and gradient boosting, have been widely used for tasks such as mortality prediction, length-of-stay estimation, and hospital readmission prediction~\cite{rajkomar2018scalable}. More recently, deep learning has emerged as a powerful paradigm in healthcare analytics. Convolutional Neural Networks (CNNs) have achieved state-of-the-art results in medical image analysis~\cite{litjens2017survey}, while Recurrent Neural Networks (RNNs), Long Short-Term Memory (LSTM) networks, and Transformer models have shown significant promise in processing sequential data like EHR time series and unstructured clinical notes~\cite{xu2023transehr}. These models excel at automatic feature extraction from high-dimensional data and learning intricate patterns that may be missed by traditional techniques. Applications span a wide range of clinical problems, including those addressed in our work such as mortality prediction, length of stay estimation, and readmission risk assessment~\cite{zhang2023transformehr}. However, these methods typically focus on a single modality, such as EHR, and struggle to incorporate complementary data from clinical notes or medical images. This reliance on single-modality inputs limits their ability to capture the full complexity of patient health, particularly in real-world scenarios where data availability varies. Furthermore, these models often require extensive feature engineering or preprocessing, reducing their scalability across diverse clinical tasks. Our MoE-Health framework addresses these limitations by dynamically integrating multiple modalities, offering a more flexible and robust solution for healthcare prediction.

\subsection{Multimodal Learning}
Multimodal learning focuses on building models that can process and relate information from multiple heterogeneous data sources~\cite{baltruvsaitis2018multimodal}. In healthcare, this involves integrating data from EHRs, clinical notes, medical images, genomics, and other physiological signals to gain a more holistic understanding of a patient's condition~\cite{zhao2024deep,krones2025review}. The central idea is that different modalities offer complementary information, and their effective fusion can lead to more accurate and robust predictions than using any single modality alone. Initially, researchers fused modality embedding vectors using methods such as direct concatenation, summation, or pooling for downstream tasks~\cite{soenksen2022integrated}. Early fusion approaches combine raw data from different modalities before processing, while late fusion combines predictions from modality-specific models. Intermediate fusion strategies, which combine features at various stages of processing, have shown superior performance in many applications. Wang et al.~\cite{wang2024multimodal} proposed a framework that integrate time-series vitals, medical images, and clinical notes with three fusion strategies. Attention-based fusion mechanisms have gained popularity for their ability to dynamically weight the importance of different modalities based on the specific prediction task and input characteristics. Yoon et al.~\cite{yoon2022d} proposed a multimodal Transformer encoder for depression detection, which incorporates two cross-attention submodules to capture inter-modal correlations. However, these approaches often assume the availability of all modalities or requiring exclusion of incomplete data, which is impractical in clinical settings where missing modalities are common. Outside healthcare, Mixture of Experts (MoE) models have shown promise in tasks like natural language processing and computer vision by leveraging specialized subnetworks~\cite{shazeer2017outrageously}. Yet, their application to multimodal healthcare data remains underexplored. MoE-Health builds on these ideas, introducing a novel MoE architecture that dynamically selects and combines modality-specific experts, ensuring robustness to missing data and improving performance on clinical tasks like those evaluated on MIMIC-IV.

\begin{figure*}[t]
  \centering
  \includegraphics[width=\textwidth]{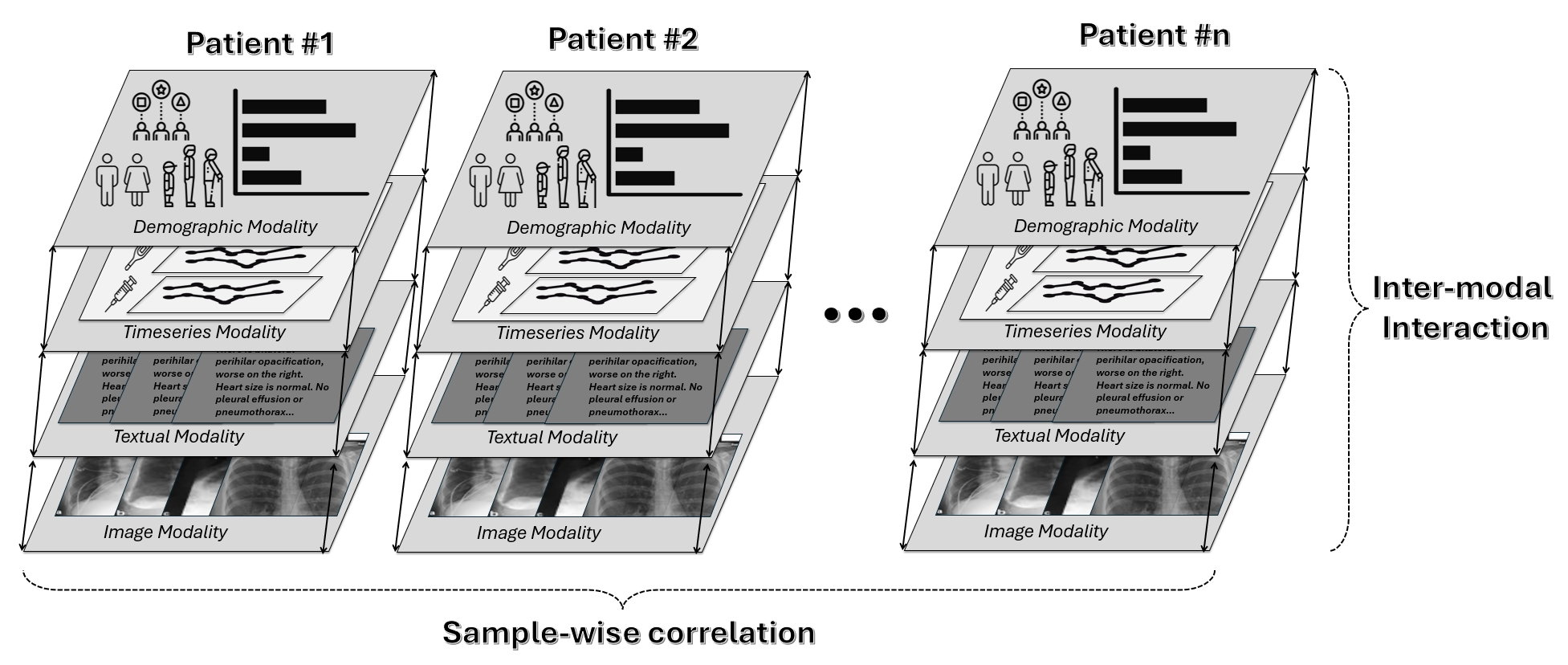}
  \caption{An overview of the multimodal patient record. Each sample integrates information from three major modalities: the EHR Modality, which encompasses the Demographic Modality (demographic attributes) and the Timeseries Modality (vital signs and laboratory test values); the Textual Modality (clinical notes); and the Image Modality (chest X-ray images).}
  \Description{Illustration of our multimodal dataset.}
  \label{fig: sample}
\end{figure*}

\section{Methodology}
\subsection{Preliminary and Notations}
\label{Preliminary_Notations}
\paragraph{Sparse Mixture-of-Experts} The Mixture-of-Experts (MoE) framework is well-suited for multimodal data due to its ability to leverage specialized expert subnetworks \( \{E_i\}_{i=1}^n \) and a gating mechanism \(G\) produces scores that guide the routing mechanism in determining which experts should process a given input~\cite{shazeer2017outrageously}. In a sparse MoE layer with \( n \) expert networks, \( E_1, E_2, \dots, E_n \), and a gating network \( G \), the gating network routes the input \( \mathbf{x} \in \mathbb{R}^d \) to a subset of \( k \) experts (\( k \ll n \)) by computing logits \( h(\mathbf{x}) = W_g \cdot \mathbf{x} \), where \( W_g \in \mathbb{R}^{n \times d} \), and selecting the top-\( k \) experts with the highest logits. The logits \( h(\mathbf{x}) \in \mathbb{R}^n \) assign a score to each expert. Let \( \mathcal{T}(\mathbf{x}) \subseteq \{1, 2, \dots, n\} \) denote the set of indices of the top-\( k \) experts for input \( \mathbf{x} \). The gating scores for the selected experts are computed by applying a softmax function over their logits:

\begin{equation}
    G(x)_i =
    \begin{cases}
    \frac{\exp(h(x)_i)}{\sum_{j \in \mathcal{T}(x)} \exp(h(x)_j)} & \text{if } i \in \mathcal{T}(x) \\\\
    0 & \text{if } i \notin \mathcal{T}(x)
\end{cases}
\end{equation}
The final output \( \mathbf{y} \in \mathbb{R}^m \) of the MoE layer is a weighted combination of the outputs from the selected experts:
\begin{equation}
    \mathbf{y} = \sum_{i \in \mathcal{T}(\mathbf{x})} G(\mathbf{x})_i \cdot E_i(\mathbf{x}),
\end{equation}
where \( E_i(\mathbf{x}) \in \mathbb{R}^m \) is the output of expert \( E_i \) for input \( \mathbf{x} \). To ensure efficient training and balanced expert utilization, additional loss terms, such as load balancing losses, are often incorporated~\cite{shazeer2017outrageously}.

\paragraph{Task 1: In-hospital Mortality Prediction} This is a binary classification task where we predict patient mortality within 48 hours following hospital admission. For sample $n$, the target is defined as:
\begin{equation}
    y_{\text{mort}}^{(n)} = 
    \begin{cases}
        1 & \text{if patient passes away within 48hrs} \cr
        0 & \text{otherwise}
    \end{cases}
\end{equation}

\paragraph{Task 2: Length of Stay Prediction} This task aims to predict whether a patient's hospital stay will be longer than 7 days, a significant outcome of interest for identifying resource-intensive hospitalizations. The prediction is made using data from the first 48 hours of admission. The label is defined as:
\begin{equation}
    y_{\text{los}} =
    \begin{cases}
    1 & \text{if length of stay } > 7 \text{ days} \\\\
    0 & \text{otherwise}
    \end{cases}
\end{equation}

\paragraph{Task 3: Hospital readmission prediction} This is a binary classification task to predict whether a patient will be readmitted within 30 days of discharge. For any patient, we utilize timeseries measurements, clinical notes, and medical images within the first 48 hours from admission to predict. The label is defined as:
\begin{equation}
    y_{\text{readmission}} =
    \begin{cases}
    1 & \text{if patient readmitted within 30 days of discharge} \\\\
    0 & \text{otherwise}
    \end{cases}
\end{equation}

\begin{figure*}[t]
  \centering
  \includegraphics[width=\textwidth]{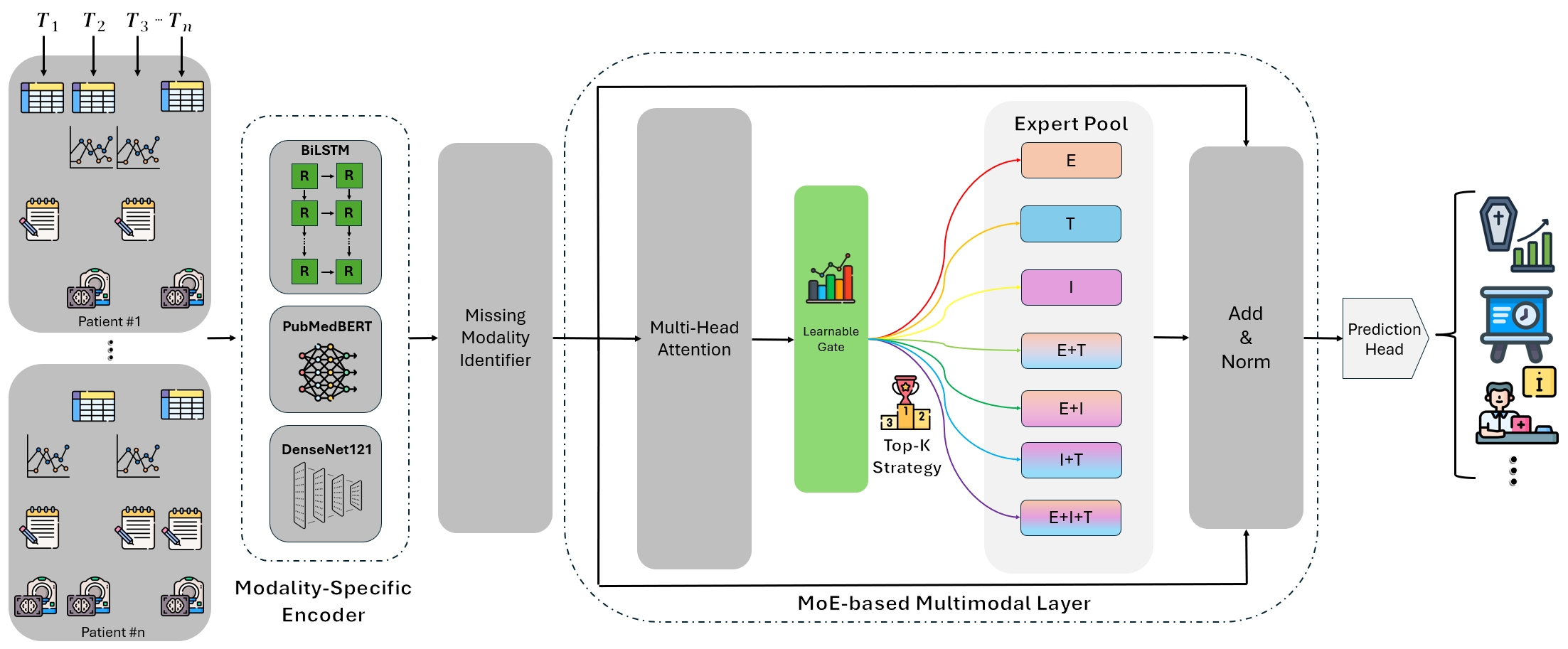}
  \caption{Overview of the proposed MoE-Health architecture. The MoE-Health is designed to flexibly process inputs with arbitrary combinations of modalities, including cases with missing data. Each modality is first encoded using a dedicated encoder, and a missing modality indicator is incorporated to inform the fusion process. At the core of the architecture is the MoE Fusion Layer, where a learned routing mechanism assigns each input to a tailored combination of expert networks. The outputs of these experts are then combined using a gating mechanism to produce a unified representation, which is passed to downstream prediction tasks.}
  \Description{Overview of the proposed MoE-Health architecture. }
  \label{fig: framework}
\end{figure*}

\subsection{The MoE-Health Architecture}
The foundational principle of our MoE-Health framework is its capacity to dynamically adapt to the inherent variability in available data modalities for each patient. The architecture is engineered to manage the complex, high-dimensional, and often sparse multimodal data characteristic of real-world clinical datasets. The process involves three main stages, as illustrated in Figure \ref{fig: framework}: (1) Modality-Specific Encoding, (2) MoE-based Multimodal Fusion, and (3) Prediction.
\subsubsection{Modality-Specific Encoding}
Our framework first encodes each modality into a fixed-dimension embedding, creating a standardized feature space. These modality-specific encoders transform raw data into high-level representations that serve as the unified input for the subsequent gating network and expert pool.
\paragraph{Feature embedding for EHR} The EHR encoder processes structured clinical data, including static demographics and dynamic time-series information. Static features, such as age, gender, and ethnicity, are encoded using normalization for numerical values and learnable embeddings for categorical variables. Dynamic time-series features, which include lab results, vital signs, and medications, are aggregated into hourly time bins within a 48-hour observation window post-admission. A bidirectional Long Short-Term Memory (BiLSTM) network~\cite{harutyunyan2019multitask} processes these sequences to capture complex temporal dependencies. The static and dynamic features are then concatenated and passed through a final linear layer to produce a comprehensive patient state embedding: 
\begin{equation}
    \mathbf{e}_{\text{EHR}} = \text{Linear}(\text{Concat}(\text{StaticEmb},\text{BiLSTM}(\mathbf{x}_{\text{EHR}}))) \in \mathbb{R}^{d_h}.
\end{equation}

\paragraph{Feature embedding for Clinical Notes} The clinical text encoder processes unstructured notes, such as radiology reports, discharge summaries, and other clinical documentation. Our preprocessing pipeline involves selecting relevant note types, removing de-identification artifacts, and extracting salient sections (e.g., "Impression," "Findings"). We then tokenize the text using a domain-specific tokenizer. For feature extraction, we employ a pre-trained ClinicalBERT model~\cite{liu2025generalist}, which is specifically trained on clinical corpora like MIMIC-III and excels at interpreting medical terminology. The final text embedding is derived from the hidden state of the [CLS] token:
\begin{equation}
    \mathbf{e}_{\text{Text}} = \text{ClinicalBERT}(\mathbf{x}_{\text{Notes}})_{\text{[CLS]}} \in \mathbb{R}^{d_h}
\end{equation}

\paragraph{Feature embedding for Images} The imaging encoder processes chest X-ray
(CXR) images from the MIMIC-CXR-JPG dataset, selecting the first relevant anteroposterior view associated with the hospital admission. Images undergo standard preprocessing, including normalization and resizing to 224x224 pixels. We utilize a DenseNet-121 architecture~\cite{huang2017densely}, pre-trained on the CheXpert dataset, as the core feature extractor. The final image embedding is extracted from the output of the penultimate layer (global average pooling) before the final classification head, capturing a rich visual representation:
\begin{equation}
    \mathbf{e}_{\text{Image}} = \text{GloablAvgPool}(\text{DenseNet121}(\mathbf{x}_{\text{CXR}})) \in \mathbb{R}^{d_h}
\end{equation}

\paragraph{Handling Modality Missingness}
To address the pervasive issue of modality incompleteness, where samples may lack data from one or more modalities, we employ a strategy based on learnable embeddings. Instead of a single shared vector, we introduce a set of modality-specific learnable indicator embeddings \( \{\mathbf{e}_{\text{absent}, m}\}_{m=1}^M \), where M is the total number of modalities and each 
$\mathbf{e}_{\text{absent}, m} \in \mathbb{R}^{d_h}$, where $d_h$ is the
unified embedding dimension. When data for the $m$-th modality is unavailable for a given sample, its corresponding encoder output is replaced by its dedicated missingness embedding, $\mathbf{e}_{\text{absent}, m}$. Each of these embeddings is initialized randomly and optimized end-to-end during training. This approach allows the model to learn a distinct and meaningful representation for the absence of each specific modality, providing a more nuanced signal to the downstream gating network and expert pool. Formally, the modality-specific embedding for sample \( i \) and modality \( m \) is defined as:
\begin{equation}
\begin{cases}
        \text{Encoder}_m(\mathbf{x}_m^{(i)}) & \text{if } m \in \mathcal{M}_i \\
\mathbf{e}_{\text{absent}, m} & \text{if } m \notin \mathcal{M}_i
\end{cases}
\end{equation}

where \( \mathcal{M}_i \) denotes the set of available modalities for sample \( i \), and \( \text{Encoder}_m(\mathbf{x}_m^{(i)}) \in \mathbb{R}^{d_h} \) is the output of the modality-specific encoder.

To form a unified multimodal representation for each sample, the individual modality embeddings are concatenated into a single flat vector. This vector aggregates the high-level features from all available sources, serving as the final input to the main network body (e.g., the MoE layer's gating and expert networks).
The concatenated multimodal representation is then constructed as:
\begin{equation}
    \mathcal{R}^{(i)} = \text{Concat}(\mathbf{v}_1^{(i)}, \mathbf{v}_2^{(i)}, \dots, \mathbf{v}_M^{(i)}) \in \mathbb{R}^{M \times  d_h},
\end{equation}
where $Concat(\cdot)$ denotes concatenation along the feature dimension.

\subsubsection{MoE-based Multimodal Fusion}
Rather than assigning individual experts to specific modalities, MoE-Health introduces a novel expert design tailored to modality combinations. This design acknowledges that the optimal fusion strategy depends on the specific subset of available modalities, and that expert specialization should be aligned with such combinations.

Let $\mathcal{C} = \{c_1, c_2, ..., c_N\}$ denote the set of distinct modality combinations observed in the training set, where $N = |\mathcal{C}|$. For each combination $c_n \in \mathcal{C}$, we instantiate a dedicated expert network $E_n$. Each expert is implemented as a lightweight multi-layer perceptron (MLP) designed to model both intra- and inter-modal interactions. To enhance expert specialization, we employ a combination-specific pretraining strategy: each $E_n$ is pre-trained exclusively on samples exhibiting combination $c_n$, allowing it to learn modality-specific co-representations before joint training.

The fusion mechanism is governed by a gating network $G(\cdot)$ that takes the concatenated multimodal embedding $\mathcal{R}^{(i)}$ as input. Missing modalities are represented by a shared learnable token $\mathbf{e}_{\text{absent}}$, ensuring that $\mathcal{R}^{(i)}$ encodes both observed data and missing patterns. The gating network, parameterized as a multi-layer perceptron (MLP), outputs a soft distribution over the $K$ experts:
\begin{equation}
    \mathbf{g}^{(i)} = G(\mathcal{R}^{(i)}) = \text{Softmax}(\mathbf{W}_g \cdot \text{ReLU}(\mathbf{W}_\mathcal{R} \mathcal{R}^{(i)} + \mathbf{b}_\mathcal{R}) + \mathbf{b}_g)
\end{equation}

where $\mathbf{g}^{(i)} \in \mathbb{R}^K$ denotes the expert routing weights for sample $i$, $\mathbf{W}_\mathcal{R} \in \mathbb{R}^{d_\mathcal{R} \times (M \times d_h)}$, and $d_\mathcal{R}$ is the hidden dimension of the MLP.

To promote diversity and robustness, we employ a top-$k$ routing strategy: only the top-$k$ experts with the highest gating scores are activated for each sample. The final output is the weighted sum of selected expert predictions:
\begin{equation}
    \mathcal{T}_k^{(i)} = \text{TopK}(\mathbf{g}^{(i)}, k), \quad
\hat{y}_i = \sum_{j \in \mathcal{T}_k^{(i)}} \frac{g_j^{(i)}}{\sum_{\ell \in \mathcal{T}_k^{(i)}} g_\ell^{(i)}} \cdot E_j(\mathcal{R}^{(i)})
\end{equation}

\paragraph{Training Objective}

The model is trained end-to-end by optimizing a composite loss function $\mathcal{L}$:

\begin{equation}
    \mathcal{L} = \mathcal{L}_{\text{task}} + \mathcal{L}_{\text{balance}}
\end{equation}

Since our predictive tasks (in-hospital mortality, length of stay, and hospital readmission) are framed as binary classification problems, we adopt the Binary Cross-Entropy (BCE) loss:
\begin{equation}
    \mathcal{L}_{\text{task}} = -\frac{1}{N} \sum_{i=1}^{N} \left[ y_i \log(\hat{y}_i) + (1 - y_i) \log(1 - \hat{y}_i) \right]
\end{equation}

To ensure expert utilization is balanced, we introduce an auxiliary load balancing loss $\mathcal{L}_{\text{balance}}$ following Shazeer et al.~\cite{shazeer2017outrageously}. Let $\mathbf{f} = [f_1, f_2, ..., f_K]$ represent the number of times each expert is selected in $\mathcal{T}_k^{(i)}$ across a batch, with:

\begin{equation}
    f_k = \sum_{i=1}^N \mathbb{I}(k \in \mathcal{T}_k^{(i)}), \quad
\mathbf{p} = \frac{1}{N} \sum_{i=1}^{N} \mathbf{g}^{(i)}
\end{equation}

Then the load balancing loss is defined as:

\begin{equation}
    \mathcal{L}_{\text{balance}} = \alpha \cdot \text{CV}(\mathbf{f} \odot \mathbf{p}), \quad \text{where} \quad \text{CV}(\cdot) = \frac{\sigma(\cdot)}{\mu(\cdot)}
\end{equation}

Here, $\odot$ denotes element-wise multiplication, and $\alpha$ is a hyperparameter controlling the balance regularization strength. When expert usage is skewed, the coefficient of variation increases, leading to a stronger regularization signal.

This modular design enables MoE-Health to dynamically route multimodal inputs based on their modality availability, robustly handle missing data, and improve performance by leveraging expert specialization across observed modality patterns.

\section{Experiments}

\subsection{Dataset}

% \begin{figure*}[t]
%   \centering
%   \includegraphics[width=\textwidth]{Distribution of Modality.png}
%   \caption{Distribution of modality availability across 31,088 hospital admissions in the utilized dataset. A: admissions with all modalities available (EHR, clinical notes, and chest X-rays); T: admissions missing clinical notes; I: admissions missing chest X-ray images; E: admissions missing EHR data (none in this dataset). The substantial proportion of admissions with missing modalities (62.6\%) highlights the critical challenge for multimodal learning in real-world clinical settings.}
%   \Description{Overview of the proposed MoE-Health architecture. }
%   \label{fig: framework}
% \end{figure*}
\begin{figure*}[t]
  \centering
  \begin{subfigure}[t]{0.48\linewidth}
    \centering
    \includegraphics[width=\linewidth]{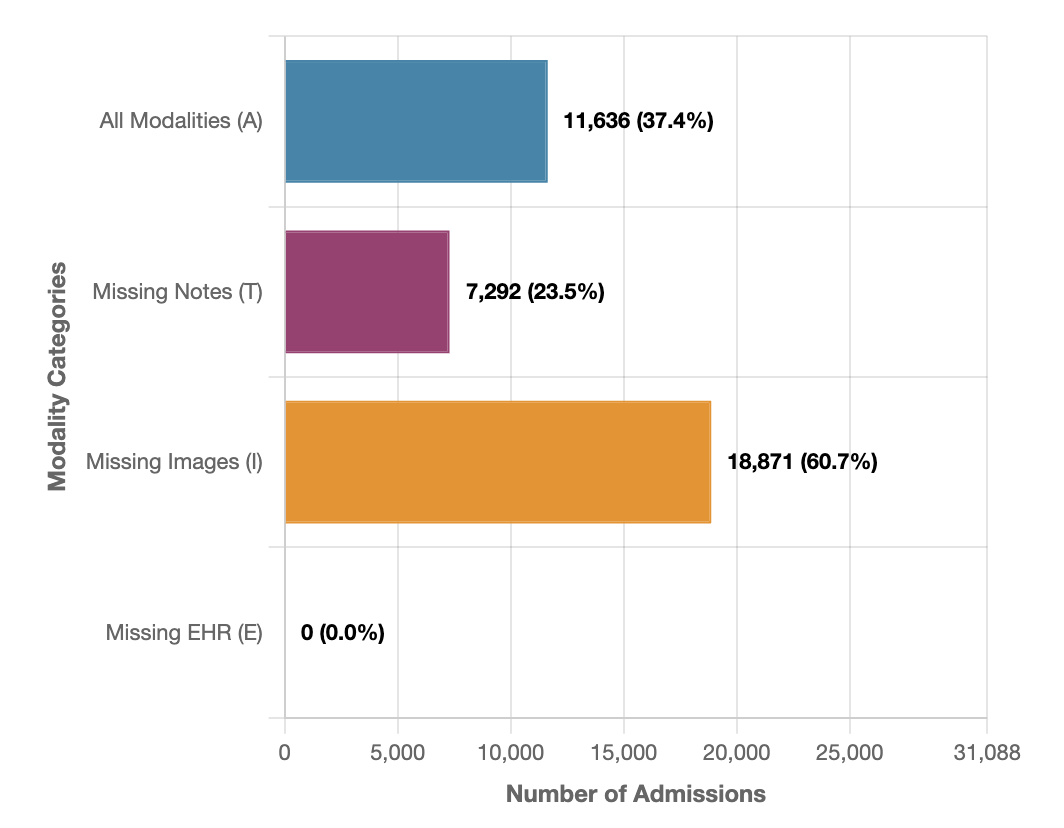}
    \caption{Distribution of modality availability across 31,088 hospital admissions in the utilized dataset. A: admissions with all modalities available (EHR, clinical notes, and chest X-rays); T: admissions missing clinical notes; I: admissions missing chest X-ray images; E: admissions missing EHR data (none in this dataset). The substantial proportion of admissions with missing modalities highlights the critical challenge for multimodal learning in real-world clinical settings.}
    \label{fig:missing_modalities}
  \end{subfigure}
  \hfill
  \begin{subfigure}[t]{0.48\linewidth}
    \centering
    \includegraphics[width=\linewidth]{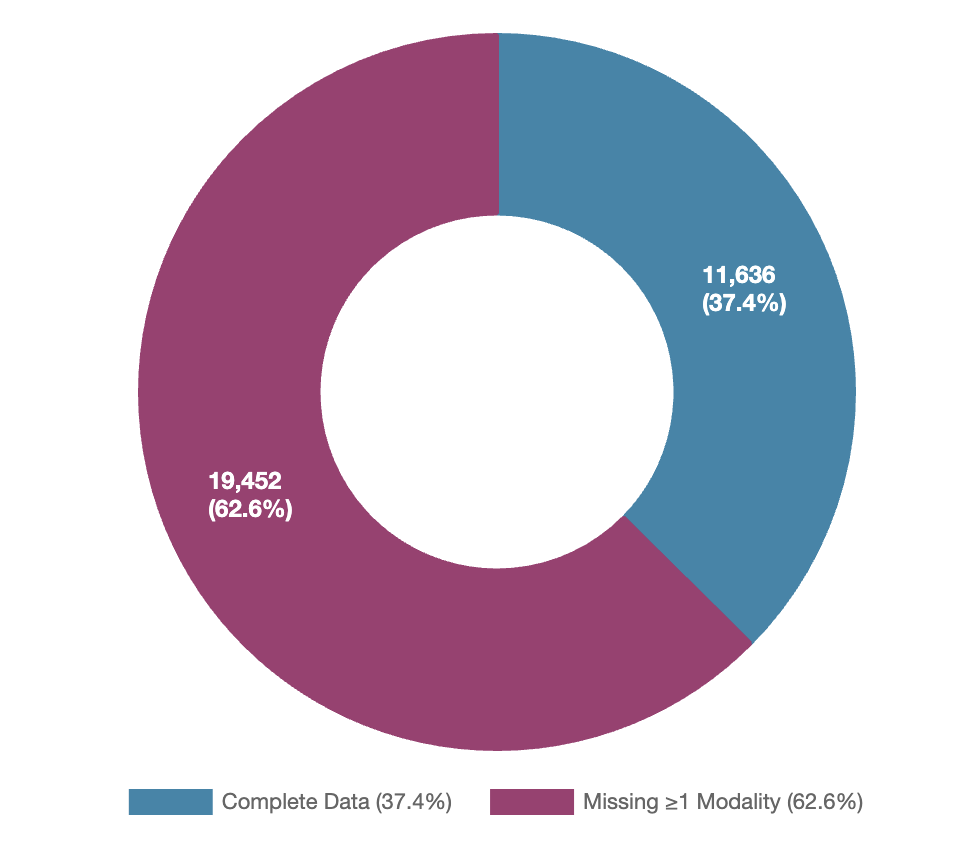}
    \caption{Overall data completeness in the MIMIC-IV dataset. Only 37.4\% of admissions have complete multimodal data available, while 62.6\% suffer from at least one missing modality. This stark contrast demonstrates the gap between ideal multimodal learning scenarios and real-world clinical data availability.}
    \label{fig:modality_combinations}
  \end{subfigure}
  \caption{Modality statistics in the MIMIC-IV dataset.}
  \label{fig:modality_overview}
\end{figure*}

% \begin{figure}[t]
%   \centering
%   \includegraphics[width=\linewidth]{Distribution of Modality.png}
%   \caption{Distribution of modality availability across 31,088 hospital admissions in the utilized dataset. A: admissions with all modalities available (EHR, clinical notes, and chest X-rays); T: admissions missing clinical notes; I: admissions missing chest X-ray images; E: admissions missing EHR data (none in this dataset). The substantial proportion of admissions with missing modalities (62.6\%) highlights the critical challenge for multimodal learning in real-world clinical settings.}
%   \Description{Distribution of modality availability across the dataset.}
%   \label{fig:modality_distribution}
% \end{figure}

\subsubsection{Dataset and Cohort Construction}

To evaluate our framework, we constructed a large-scale multimodal dataset by integrating two publicly available resources: the Medical Information Mart for Intensive Care (MIMIC-IV) v2.2 and the MIMIC-CXR-JPG v2.0.0 dataset~\cite{johnson2024mimic, johnson2019mimic}. Our integration strategy is conceptually aligned with the methodology proposed by Soenksen et al.~\cite{soenksen2022integrated}. MIMIC-IV is a comprehensive, de-identified clinical database that includes the electronic health records (EHRs) of nearly 300,000 patients admitted to the Beth Israel Deaconess Medical Center (BIDMC) between 2008 and 2019. We focus on a subset of 73,181 ICU stays, from which we extracted three data modalities: structured EHR data (comprising static demographic variables and dynamic time-series signals), unstructured clinical notes, and imaging metadata. MIMIC-CXR-JPG contains 377,110 chest radiographs in JPEG format, corresponding to 227,835 imaging studies. This dataset is directly linkable to MIMIC-IV through anonymized patient identifiers and imaging study identifiers, enabling multimodal alignment across hospital stays. Using these identifiers, we derived the final cohort for our experiments, as illustrated in Fig.~\ref{fig: sample}. The resulting dataset includes 31,088 unique hospital admissions with at least one available modality among EHR, chest X-ray (CXR), and clinical notes. A key characteristic of this cohort—reflecting real-world clinical conditions—is the heterogeneous availability of modalities across patients. As shown in Fig.~\ref{fig:modality_overview}, EHR data is available for all admissions (100\%), while only 11,636 admissions (37.4\%) contain all three modalities. CXR images are missing in 18,871 cases (60.7\%), and clinical notes are missing in 7,292 cases (23.5\%). This high degree of missingness, stemming from varying clinical documentation and imaging practices, highlights the need for models that can robustly operate under partial modality settings. Accordingly, this dataset serves as an ideal testbed for evaluating the robustness and flexibility of our proposed MoE-Health framework, which is specifically designed to accommodate arbitrary combinations of available modalities.

%For more detailed data table with the demographic characteristics, laboratory tests, and labels of the study samples, please refer to Appendix~\ref{appendix: dataset}.

\subsection{Baselines}
To comprehensively evaluate the performance of our proposed MoE-Health framework, we compare it against a variety of baseline models. These baselines include unimodal models to establish single-modality performance levels, traditional fusion methods, and advanced multimodal architectures.
\paragraph{Unimodal Baselines} These models are trained using data from a single modality to provide a performance benchmark for each data type individually. The bidirectional LSTM for timeseries EHR only, pre-trained DenseNet-121 for CXR images only, and pre-trained ClinicalBERT for clinical notes are included for comparison. Those encoders are connected directly to a prediction head.
\paragraph{Traditional Multimodal Fusion Methods} Those methods represent common, non-dynamic approaches to multimodal fusion, including Early Fusion, Late Fusion (Ensemble), and Joint Fusion. MPP~\cite{wang2024multimodal} is a tri-modal fusion model that applied three mentioned fusion strategies to integrate time series EHR, medical images, and clinical notes. We also incorporated the HAIM method~\cite{soenksen2022integrated}, a data pipeline specifically designed for integrating multimodal data from the MIMIC-IV dataset. 
\paragraph{Advanced Multimodal Baselines}
We also compare against more sophisticated multimodal fusion architectures from recent literature to benchmark against the state-of-the-art. TriMF~\cite{wang2025missing} is a framework that first fuses two modalities each and then fuses the resulting three BiMF representations together.

All baselines were trained or fine-tuned on the same dataset splits and tasks as MoE-Health. Notably, MPP and TriMF were designed to be trained and tested on samples with complete modalities. 
\subsection{Experimental Settings}
We evaluate all models on the three binary classification tasks defined in Section \ref{Preliminary_Notations}: in-hospital mortality,  length of stay, and hospital readmission. To assess model performance comprehensively, we use two standard evaluation metrics: Area Under the Receiver Operating Characteristic Curve (AUROC) and F1-Score. The final patient dataset is split into training, validation, and test sets using a ratio of 80:10:10 for model training and evaluation.

All models are implemented using the PyTorch framework. Training is conducted using the AdamW optimizer with a learning rate of 0.0001 and a batch size of 32. Models are trained for a maximum of 50 epochs with an early stopping mechanism based on the validation set's AUROC score to prevent overfitting. All experiments are run on an NVIDIA A40 GPU. The scaling coefficient $\alpha$ for the auxiliary loss is set to 0.01. The number for the top-k strategy is 2, and the number of attention heads is 4.

\subsection{Results}
\subsubsection{Results of Comparative Experiments}
\begin{table*}[!htbp]
\caption{Comparison of our proposed methods (MoE-Health) and baselines on three binary prediction tasks. The best results are highlighted in bold font.}
\label{tab:comparison}
\begin{tabular}{cccccccc}
\toprule
Task / Methods                                              &                               & HAIM           & MPP-Early & MPP-Joint & MPP-Late & TriMF          & MoE-Health     \\ \hline
\multicolumn{1}{c|}{\multirow{2}{*}{In-hospital Mortality}} & \multicolumn{1}{l|}{AUROC}    & 0.783          & 0.790     & 0.800     & 0.797    & 0.806          & \textbf{0.818} \\
\multicolumn{1}{c|}{}                                       & \multicolumn{1}{l|}{F1 Score} & 0.398          & 0.406     & 0.425     & 0.423    & 0.435          & \textbf{0.465} \\ \hline
\multicolumn{1}{l|}{\multirow{2}{*}{Long Length of Stay}}   & \multicolumn{1}{l|}{AUROC}    & 0.782          & 0.748     & 0.775     & 0.763    & 0.771          & \textbf{0.794} \\
\multicolumn{1}{l|}{}                                       & \multicolumn{1}{l|}{F1 Score} & \textbf{0.739} & 0.724     & 0.733     & 0.729    & 0.728          & \textbf{0.739} \\ \hline
\multicolumn{1}{l|}{\multirow{2}{*}{Hospital Readmission}}  & \multicolumn{1}{l|}{AUROC}    & 0.620          & 0.616     & 0.637     & 0.623    & 0.638          & \textbf{0.643} \\
\multicolumn{1}{l|}{}                                       & \multicolumn{1}{l|}{F1 Score} & 0.286          & 0.278     & 0.285     & 0.269    & \textbf{0.289} & 0.281          \\ \bottomrule
\end{tabular}
\end{table*}

We present comprehensive comparative analysis results across three prediction tasks, evaluating our proposed MoE-Health framework against established multimodal fusion baselines. Table \ref{tab:comparison} summarizes the performance comparison using AUROC and F1-score metrics. Our MoE-Health framework demonstrates superior performance across the majority of evaluation metrics and prediction tasks. Specifically, MoE-Health achieves the highest AUROC scores for all three prediction tasks: 0.818 for in-hospital mortality prediction (improvement of 1.2 percentage points over the second-best TriMF method), 0.794 for long length of stay prediction (improvement of 1.2 percentage points over HAIM), and 0.643 for hospital readmission prediction (improvement of 0.5 percentage points over TriMF).
For F1-score performance, MoE-Health achieves the best results in two out of three tasks. In in-hospital mortality prediction, our method obtains an F1-score of 0.465, representing a substantial improvement of 3.0 percentage points over the second-best TriMF method (0.435). For long length of stay prediction, MoE-Health ties with HAIM at 0.739 F1-score, both outperforming other baseline methods. In hospital readmission prediction, while TriMF achieves the highest F1-score (0.289), MoE-Health demonstrates competitive performance at 0.281.

% \begin{table}[H]
% \caption{Performance of MoE-Health and TriMF on In-hospital Mortality across different modality combinations, measured by AUROC and F1 Score. Best results are \textbf{bolded}.}
% \label{tab:modality}
% \centering
% \begin{tabular}{lccccc}
% \toprule
% Modality Combination & \multicolumn{2}{c}{MoE-Health} & \multicolumn{2}{c}{TriMF} \\
% \cmidrule(lr){2-3} \cmidrule(lr){4-5}
% & AUROC & F1 Score & AUROC & F1 Score \\ \hline
% EHR & \textbf{0.760} & \textbf{0.410} & 0.730 & 0.390 \\
% IMG & \textbf{0.740} & \textbf{0.400} & 0.710 & 0.380 \\
% TXT & \textbf{0.750} & \textbf{0.405} & 0.720 & 0.385 \\
% EHR+IMG & \textbf{0.790} & \textbf{0.435} & 0.770 & 0.415 \\
% EHR+TXT & \textbf{0.785} & \textbf{0.440} & 0.765 & 0.420 \\
% IMG+TXT & \textbf{0.780} & \textbf{0.430} & 0.760 & 0.410 \\
% EHR+IMG+TXT & \textbf{0.818} & \textbf{0.465} & 0.806 & 0.435 \\ \bottomrule
% \end{tabular}
% \end{table}
\subsubsection{Results of Modality Combination Analysis}
To further investigate the synergistic effects of multimodal data fusion, we evaluated the performance of our MoE-Health framework using different combinations of the available modalities: EHR (E), Clinical Notes (T), and Image (I). The results for the Long Length of Stay (LOS) prediction task are presented in Table \ref{tab:modality}. 

\begin{table}[H]
\caption{Performance of MoE-Health on In-hospital Mortality across different modality combinations, measured by AUROC and F1 Score. Best results are \textbf{bolded}.}
\label{tab:modality}
\begin{tabular}{lcc}
\toprule
\textbf{Modality Combination} & \textbf{AUROC} & \textbf{F1 Score} \\ \hline
\textbf{Single Modality     }          &                &                   \\ \hline
EHR (E)                       & 0.770          & 0.725             \\
Text (T)                      & 0.725          & 0.690             \\
Image (I)                     & 0.701          & 0.670             \\ \hline
\textbf{Dual Modality}                 &                &                   \\ \hline
E + T                         & 0.785          & 0.732             \\
E + I                         & 0.781          & 0.730             \\
T + I                         & 0.740          & 0.705             \\ \hline
\textbf{Full Modality}                 &                &                   \\ \hline
E + T + I                     & \textbf{0.794}          & \textbf{0.739}            \\ \bottomrule
\end{tabular}
\end{table}
The results clearly demonstrate the value of data fusion and highlight the relative importance of each modality for this specific task. Among the single modalities, EHR data provides the strongest predictive signal, achieving an AUROC of 0.770 and an F1 score of 0.725, establishing it as the foundational data source. Clinical notes (Text) also offer significant predictive power, while imaging data alone is the least informative for LOS prediction. Combining modalities consistently leads to performance improvements across both metrics. The addition of either Text or Image data to the EHR modality (E+T and E+I) boosts the performance, with the combination of EHR and Text yielding the best dual-modality results (AUROC 0.785, F1 0.732). This shows that both unstructured notes and medical images provide complementary information that is not fully captured in the structured EHR data. Crucially, the combination of all three modalities (E+T+I) yields the highest performance, achieving an AUROC of 0.794 and an F1 score of 0.739. This result confirms that integrating all available data sources allows the model to form the most comprehensive and accurate understanding of a patient's condition. The incremental gains observed with each added modality underscore the effectiveness of our fusion architecture in leveraging these diverse data streams.

\subsubsection{Results of Ablation Experiments}
To dissect the architecture of MoE-Health and quantify the contribution of its key design elements, we conducted a series of ablation studies. We systematically removed or replaced individual components from the full model and evaluated the performance on the in-hospital mortality task. The results, presented in Table \ref{tab:ablation}, highlight the impact of each component on the model's overall efficacy.

\begin{table}[H]
\caption{Ablation Study Results (AUROC for In-Hospital Mortality)}
\label{tab:ablation}
\begin{tabular}{lcc}
\toprule
Configuration             & AUROC & $\Delta$ AUROC \\ \hline
MoE-Health (Full)         & 0.818 & -       \\ \hline
w/o Missing Indicator     & 0.788 & -0.030  \\
w/o Expert Specialization & 0.735 & -0.083  \\
w/o Dynamic Gating        & 0.765 & -0.053  \\
w/o Top-k Routing         & 0.801 & -0.017  \\ \bottomrule
\end{tabular}
\end{table}

The analysis reveals that Expert Specialization is the most critical component of our framework. Removing the specialized, pre-trained experts and replacing them with generic, non-specialized ones resulted in the most substantial performance degradation, with a drop in AUROC of 0.083. This finding strongly supports our hypothesis that dedicating experts to specific modality combinations is essential for effective fusion. The second most impactful component is Dynamic Gating. When the learnable gating network was replaced with a non-dynamic mechanism (e.g., simple averaging of all expert outputs), the AUROC decreased by 0.053. This underscores the importance of a data-driven routing strategy that can intelligently assign inputs to the most appropriate experts. Furthermore, removing the learnable Missing Indicator vector and using simple zero-padding for absent modalities led to a performance drop of 0.030 in AUROC. This indicates that having an explicit, trainable representation to signal missingness provides a valuable signal to the model.
Finally, simplifying the routing strategy from Top-k (k=2) to Top-1 (i.e., w/o Top-k Routing) resulted in a smaller but still significant drop of 0.017. This suggests that incorporating a secondary "backup" expert adds a layer of robustness to the final prediction.

In summary, the ablation study confirms that all major components of the MoE-Health architecture contribute positively to its performance, with the modality-specific expert specialization and dynamic gating mechanism being the most influential factors.

\section{Discussion}
In this study, we proposed MoE-Health, a novel Mixture of Experts framework designed to address the challenge of robust prediction from incomplete and heterogeneous multimodal healthcare data. Our experimental results demonstrate the effectiveness of this approach in a setting that closely mirrors real-world clinical scenarios. The clinical decision-making process is inherently multimodal; physicians synthesize information from a patient's dynamic physiological trends (time-series data), narrative reports (text), and diagnostic images to form a holistic assessment~\cite{lyu2023multimodal}. Our work underscores the importance of developing AI models that can mimic this complex integration.

Many existing multimodal methods, while innovative, often simplify the clinical reality. Some approaches rely heavily on static data, overlooking the crucial temporal dynamics captured in time-series EHR data that reflect a patient's evolving condition~\cite{wang2025missing}. Others are developed and validated on curated datasets where all samples are required to have a complete set of modalities~\cite{wang2024multimodal,wang2024multimodalfusion}. This "complete-case" assumption severely limits their applicability in real-world hospital environments, where data is frequently missing. Our framework directly confronts this issue by design. The comparative analysis (Table \ref{tab:comparison}) shows that MoE-Health consistently achieves superior or highly competitive performance, particularly in terms of AUROC, when compared to a range of established baseline methods. This highlights its strong discriminative power in a realistic, incomplete data setting.

Our modality combination analysis (Table \ref{tab:modality}) reinforces the value of multimodal fusion. While structured EHR data serves as the most powerful single predictor for long length of stay, the incremental performance gains from adding text and imaging data confirm that these modalities offer complementary, non-redundant information. The fact that the full three-modality combination yielded the best results validates the principle of holistic data integration, which our MoE-Health framework is designed to facilitate.

The core strength of our framework lies in its architectural design, which was validated by our ablation studies (Table \ref{tab:ablation}). The most significant finding is that Expert Specialization—pre-training experts on specific modality combinations—is the single most crucial component for model performance. This confirms our central hypothesis that dedicating expert pathways to handle different data availability scenarios is more effective than using a single, monolithic fusion mechanism. The substantial performance drop observed upon removing the Dynamic Gating and Missing Indicator components further underscores the importance of a data-driven routing mechanism and an explicit representation of missingness.

Despite the promising results, there are plenty of future directions to be explored. We plan to evaluate the generalizability of the MoE-Health framework on datasets from different institutions. We also aim to extend the architecture to incorporate additional data modalities, such as genomics or different imaging modalities (e.g., CT scans), which could provide an even more comprehensive patient view and are a promising avenue for future work. A key research direction will be to enhance the model's interpretability, for instance, by analyzing the gating network's routing decisions to understand which experts are activated for different patient profiles, potentially providing clinicians with deeper insights into the model's reasoning process.

\section{Conclusion}
In this paper, we introduced MoE-Health, a novel Mixture of Experts framework designed to perform robust clinical outcome prediction using heterogeneous and incomplete multimodal data. By leveraging modality-specific encoders and a dynamic MoE-based fusion layer with specialized experts, our model effectively adapts to the varied data availability inherent in real-world clinical settings.

Our comprehensive evaluation on the MIMIC-IV dataset demonstrates that MoE-Health achieves state-of-the-art or highly competitive performance on three challenging prediction tasks: in-hospital mortality, length of stay, and hospital readmission. The framework's success is attributed to its ability to dynamically route information to experts specialized in handling specific combinations of available modalities.

We conclude that the MoE-Health framework offers a flexible, robust, and high-performing solution to the persistent challenge of multimodal data fusion in healthcare. This work represents a significant step towards the development of more powerful and clinically applicable machine learning models that can effectively harness the full spectrum of patient data.

\bibliographystyle{ACM-Reference-Format}
\bibliography{main}


\begin{thebibliography}{22}

%%% the Web address.

\ifx \showCODEN    \undefined \def \showCODEN     #1{\unskip}     \fi
\ifx \showISBNx    \undefined \def \showISBNx     #1{\unskip}     \fi
\ifx \showISBNxiii \undefined \def \showISBNxiii  #1{\unskip}     \fi
\ifx \showISSN     \undefined \def \showISSN      #1{\unskip}     \fi
\ifx \showLCCN     \undefined \def \showLCCN      #1{\unskip}     \fi
\ifx \shownote     \undefined \def \shownote      #1{#1}          \fi
\ifx \showarticletitle \undefined \def \showarticletitle #1{#1}   \fi
\ifx \showURL      \undefined \def \showURL       {\relax}        \fi
% invisible to TeX
\providecommand\bibfield[2]{#2}
\providecommand\bibinfo[2]{#2}
\providecommand\natexlab[1]{#1}
\providecommand\showeprint[2][]{arXiv:#2}

\bibitem[Arnaud et~al\mbox{.}(2023)]%
        {arnaud2023predictive}
\bibfield{author}{\bibinfo{person}{Emilien Arnaud}, \bibinfo{person}{Mahmoud Elbattah}, \bibinfo{person}{Christine Ammirati}, \bibinfo{person}{Gilles Dequen}, {and} \bibinfo{person}{Daniel~Aiham Ghazali}.} \bibinfo{year}{2023}\natexlab{}.
\newblock \showarticletitle{Predictive models in emergency medicine and their missing data strategies: a systematic review}.
\newblock \bibinfo{journal}{\emph{NPJ Digital Medicine}} \bibinfo{volume}{6}, \bibinfo{number}{1} (\bibinfo{year}{2023}), \bibinfo{pages}{28}.
\newblock


\bibitem[Baltru{\v{s}}aitis et~al\mbox{.}(2018)]%
        {baltruvsaitis2018multimodal}
\bibfield{author}{\bibinfo{person}{Tadas Baltru{\v{s}}aitis}, \bibinfo{person}{Chaitanya Ahuja}, {and} \bibinfo{person}{Louis-Philippe Morency}.} \bibinfo{year}{2018}\natexlab{}.
\newblock \showarticletitle{Multimodal machine learning: A survey and taxonomy}.
\newblock \bibinfo{journal}{\emph{IEEE transactions on pattern analysis and machine intelligence}} \bibinfo{volume}{41}, \bibinfo{number}{2} (\bibinfo{year}{2018}), \bibinfo{pages}{423--443}.
\newblock


\bibitem[Harutyunyan et~al\mbox{.}(2019)]%
        {harutyunyan2019multitask}
\bibfield{author}{\bibinfo{person}{Hrayr Harutyunyan}, \bibinfo{person}{Hrant Khachatrian}, \bibinfo{person}{David~C Kale}, \bibinfo{person}{Greg Ver~Steeg}, {and} \bibinfo{person}{Aram Galstyan}.} \bibinfo{year}{2019}\natexlab{}.
\newblock \showarticletitle{Multitask learning and benchmarking with clinical time series data}.
\newblock \bibinfo{journal}{\emph{Scientific data}} \bibinfo{volume}{6}, \bibinfo{number}{1} (\bibinfo{year}{2019}), \bibinfo{pages}{96}.
\newblock


\bibitem[Huang et~al\mbox{.}(2017)]%
        {huang2017densely}
\bibfield{author}{\bibinfo{person}{Gao Huang}, \bibinfo{person}{Zhuang Liu}, \bibinfo{person}{Laurens Van Der~Maaten}, {and} \bibinfo{person}{Kilian~Q Weinberger}.} \bibinfo{year}{2017}\natexlab{}.
\newblock \showarticletitle{Densely connected convolutional networks}. In \bibinfo{booktitle}{\emph{Proceedings of the IEEE conference on computer vision and pattern recognition}}. \bibinfo{pages}{4700--4708}.
\newblock


\bibitem[Johnson et~al\mbox{.}({[n.\,d.]})]%
        {johnson2024mimic}
\bibfield{author}{\bibinfo{person}{Alistair Johnson}, \bibinfo{person}{Lucas Bulgarelli}, \bibinfo{person}{Tom Pollard}, \bibinfo{person}{Steven Horng}, \bibinfo{person}{Leo~Anthony Celi}, {and} \bibinfo{person}{Roger Mark}.} \bibinfo{year}{[n.\,d.]}\natexlab{}.
\newblock \bibinfo{title}{{MIMIC}-{IV}}.
\newblock
\href{https://doi.org/10.13026/6MM1-EK67}{doi:\nolinkurl{10.13026/6MM1-EK67}}


\bibitem[Johnson et~al\mbox{.}(2019)]%
        {johnson2019mimic}
\bibfield{author}{\bibinfo{person}{Alistair~EW Johnson}, \bibinfo{person}{Tom~J Pollard}, \bibinfo{person}{Seth~J Berkowitz}, \bibinfo{person}{Nathaniel~R Greenbaum}, \bibinfo{person}{Matthew~P Lungren}, \bibinfo{person}{Chih-ying Deng}, \bibinfo{person}{Roger~G Mark}, {and} \bibinfo{person}{Steven Horng}.} \bibinfo{year}{2019}\natexlab{}.
\newblock \showarticletitle{MIMIC-CXR, a de-identified publicly available database of chest radiographs with free-text reports}.
\newblock \bibinfo{journal}{\emph{Scientific data}} \bibinfo{volume}{6}, \bibinfo{number}{1} (\bibinfo{year}{2019}), \bibinfo{pages}{317}.
\newblock


\bibitem[Krones et~al\mbox{.}(2025)]%
        {krones2025review}
\bibfield{author}{\bibinfo{person}{Felix Krones}, \bibinfo{person}{Umar Marikkar}, \bibinfo{person}{Guy Parsons}, \bibinfo{person}{Adam Szmul}, {and} \bibinfo{person}{Adam Mahdi}.} \bibinfo{year}{2025}\natexlab{}.
\newblock \showarticletitle{Review of multimodal machine learning approaches in healthcare}.
\newblock \bibinfo{journal}{\emph{Information Fusion}}  \bibinfo{volume}{114} (\bibinfo{year}{2025}), \bibinfo{pages}{102690}.
\newblock


\bibitem[Litjens et~al\mbox{.}(2017)]%
        {litjens2017survey}
\bibfield{author}{\bibinfo{person}{Geert Litjens}, \bibinfo{person}{Thijs Kooi}, \bibinfo{person}{Babak~Ehteshami Bejnordi}, \bibinfo{person}{Arnaud Arindra~Adiyoso Setio}, \bibinfo{person}{Francesco Ciompi}, \bibinfo{person}{Mohsen Ghafoorian}, \bibinfo{person}{Jeroen~AW van~der Laak}, \bibinfo{person}{Bram van Ginneken}, {and} \bibinfo{person}{Clara~I S{\'a}nchez}.} \bibinfo{year}{2017}\natexlab{}.
\newblock \showarticletitle{A survey on deep learning in medical image analysis}.
\newblock \bibinfo{journal}{\emph{Medical image analysis}}  \bibinfo{volume}{42} (\bibinfo{year}{2017}), \bibinfo{pages}{60--88}.
\newblock


\bibitem[Liu et~al\mbox{.}(2025)]%
        {liu2025generalist}
\bibfield{author}{\bibinfo{person}{Xiaohong Liu}, \bibinfo{person}{Hao Liu}, \bibinfo{person}{Guoxing Yang}, \bibinfo{person}{Zeyu Jiang}, \bibinfo{person}{Shuguang Cui}, \bibinfo{person}{Zhaoze Zhang}, \bibinfo{person}{Huan Wang}, \bibinfo{person}{Liyuan Tao}, \bibinfo{person}{Yongchang Sun}, \bibinfo{person}{Zhu Song}, {et~al\mbox{.}}} \bibinfo{year}{2025}\natexlab{}.
\newblock \showarticletitle{A generalist medical language model for disease diagnosis assistance}.
\newblock \bibinfo{journal}{\emph{Nature Medicine}} (\bibinfo{year}{2025}), \bibinfo{pages}{1--11}.
\newblock


\bibitem[Lyu et~al\mbox{.}(2023)]%
        {lyu2023multimodal}
\bibfield{author}{\bibinfo{person}{Weimin Lyu}, \bibinfo{person}{Xinyu Dong}, \bibinfo{person}{Rachel Wong}, \bibinfo{person}{Songzhu Zheng}, \bibinfo{person}{Kayley Abell-Hart}, \bibinfo{person}{Fusheng Wang}, {and} \bibinfo{person}{Chao Chen}.} \bibinfo{year}{2023}\natexlab{}.
\newblock \showarticletitle{A multimodal transformer: Fusing clinical notes with structured ehr data for interpretable in-hospital mortality prediction}. In \bibinfo{booktitle}{\emph{AMIA Annual Symposium Proceedings}}, Vol.~\bibinfo{volume}{2022}. \bibinfo{pages}{719}.
\newblock


\bibitem[Rajkomar et~al\mbox{.}(2018)]%
        {rajkomar2018scalable}
\bibfield{author}{\bibinfo{person}{Alvin Rajkomar}, \bibinfo{person}{Eyal Oren}, \bibinfo{person}{Kai Chen}, \bibinfo{person}{Andrew~M Dai}, \bibinfo{person}{Nissan Hajaj}, \bibinfo{person}{Michaela Hardt}, \bibinfo{person}{Peter~J Liu}, \bibinfo{person}{Xiaobing Liu}, \bibinfo{person}{Jake Marcus}, \bibinfo{person}{Mimi Sun}, {et~al\mbox{.}}} \bibinfo{year}{2018}\natexlab{}.
\newblock \showarticletitle{Scalable and accurate deep learning with electronic health records}.
\newblock \bibinfo{journal}{\emph{NPJ digital medicine}} \bibinfo{volume}{1}, \bibinfo{number}{1} (\bibinfo{year}{2018}), \bibinfo{pages}{18}.
\newblock


\bibitem[Shaik et~al\mbox{.}(2024)]%
        {shaik2024survey}
\bibfield{author}{\bibinfo{person}{Thanveer Shaik}, \bibinfo{person}{Xiaohui Tao}, \bibinfo{person}{Lin Li}, \bibinfo{person}{Haoran Xie}, {and} \bibinfo{person}{Juan~D Vel{\'a}squez}.} \bibinfo{year}{2024}\natexlab{}.
\newblock \showarticletitle{A survey of multimodal information fusion for smart healthcare: Mapping the journey from data to wisdom}.
\newblock \bibinfo{journal}{\emph{Information Fusion}}  \bibinfo{volume}{102} (\bibinfo{year}{2024}), \bibinfo{pages}{102040}.
\newblock


\bibitem[Shazeer et~al\mbox{.}(2017)]%
        {shazeer2017outrageously}
\bibfield{author}{\bibinfo{person}{Noam Shazeer}, \bibinfo{person}{Azalia Mirhoseini}, \bibinfo{person}{Krzysztof Maziarz}, \bibinfo{person}{Andy Davis}, \bibinfo{person}{Quoc Le}, \bibinfo{person}{Geoffrey Hinton}, {and} \bibinfo{person}{Jeff Dean}.} \bibinfo{year}{2017}\natexlab{}.
\newblock \showarticletitle{Outrageously large neural networks: The sparsely-gated mixture-of-experts layer}.
\newblock \bibinfo{journal}{\emph{arXiv preprint arXiv:1701.06538}} (\bibinfo{year}{2017}).
\newblock


\bibitem[Soenksen et~al\mbox{.}(2022)]%
        {soenksen2022integrated}
\bibfield{author}{\bibinfo{person}{Luis~R Soenksen}, \bibinfo{person}{Yu Ma}, \bibinfo{person}{Cynthia Zeng}, \bibinfo{person}{Leonard Boussioux}, \bibinfo{person}{Kimberly Villalobos~Carballo}, \bibinfo{person}{Liangyuan Na}, \bibinfo{person}{Holly~M Wiberg}, \bibinfo{person}{Michael~L Li}, \bibinfo{person}{Ignacio Fuentes}, {and} \bibinfo{person}{Dimitris Bertsimas}.} \bibinfo{year}{2022}\natexlab{}.
\newblock \showarticletitle{Integrated multimodal artificial intelligence framework for healthcare applications}.
\newblock \bibinfo{journal}{\emph{NPJ digital medicine}} \bibinfo{volume}{5}, \bibinfo{number}{1} (\bibinfo{year}{2022}), \bibinfo{pages}{149}.
\newblock


\bibitem[Wang et~al\mbox{.}(2024a)]%
        {wang2024multimodalfusion}
\bibfield{author}{\bibinfo{person}{Chutong Wang}, \bibinfo{person}{Xuebing Yang}, \bibinfo{person}{Mengxuan Sun}, \bibinfo{person}{Yifan Gu}, \bibinfo{person}{Jinghao Niu}, {and} \bibinfo{person}{Wensheng Zhang}.} \bibinfo{year}{2024}\natexlab{a}.
\newblock \showarticletitle{Multimodal fusion network for ICU patient outcome prediction}.
\newblock \bibinfo{journal}{\emph{Neural Networks}}  \bibinfo{volume}{180} (\bibinfo{year}{2024}), \bibinfo{pages}{106672}.
\newblock


\bibitem[Wang et~al\mbox{.}(2025)]%
        {wang2025missing}
\bibfield{author}{\bibinfo{person}{Muyu Wang}, \bibinfo{person}{Shiyu Fan}, \bibinfo{person}{Yichen Li}, \bibinfo{person}{Zhongrang Xie}, {and} \bibinfo{person}{Hui Chen}.} \bibinfo{year}{2025}\natexlab{}.
\newblock \showarticletitle{Missing-modality enabled multi-modal fusion architecture for medical data}.
\newblock \bibinfo{journal}{\emph{Journal of Biomedical Informatics}}  \bibinfo{volume}{164} (\bibinfo{year}{2025}), \bibinfo{pages}{104796}.
\newblock


\bibitem[Wang et~al\mbox{.}(2024b)]%
        {wang2024multimodal}
\bibfield{author}{\bibinfo{person}{Yuanlong Wang}, \bibinfo{person}{Changchang Yin}, {and} \bibinfo{person}{Ping Zhang}.} \bibinfo{year}{2024}\natexlab{b}.
\newblock \showarticletitle{Multimodal risk prediction with physiological signals, medical images and clinical notes}.
\newblock \bibinfo{journal}{\emph{Heliyon}} \bibinfo{volume}{10}, \bibinfo{number}{5} (\bibinfo{year}{2024}).
\newblock


\bibitem[Xu et~al\mbox{.}(2023)]%
        {xu2023transehr}
\bibfield{author}{\bibinfo{person}{Yanbo Xu}, \bibinfo{person}{Shangqing Xu}, \bibinfo{person}{Manav Ramprassad}, \bibinfo{person}{Alexey Tumanov}, {and} \bibinfo{person}{Chao Zhang}.} \bibinfo{year}{2023}\natexlab{}.
\newblock \showarticletitle{TransEHR: Self-Supervised Transformer for Clinical Time Series Data}. In \bibinfo{booktitle}{\emph{Proceedings of the 3rd Machine Learning for Health Symposium}}. PMLR, \bibinfo{pages}{623--635}.
\newblock


\bibitem[Yoon et~al\mbox{.}(2022)]%
        {yoon2022d}
\bibfield{author}{\bibinfo{person}{Jeewoo Yoon}, \bibinfo{person}{Chaewon Kang}, \bibinfo{person}{Seungbae Kim}, {and} \bibinfo{person}{Jinyoung Han}.} \bibinfo{year}{2022}\natexlab{}.
\newblock \showarticletitle{D-vlog: Multimodal vlog dataset for depression detection}. In \bibinfo{booktitle}{\emph{Proceedings of the AAAI Conference on Artificial Intelligence}}, Vol.~\bibinfo{volume}{36}. \bibinfo{pages}{12226--12234}.
\newblock


\bibitem[Zhang et~al\mbox{.}(2022)]%
        {zhang2022m3care}
\bibfield{author}{\bibinfo{person}{Chaohe Zhang}, \bibinfo{person}{Xu Chu}, \bibinfo{person}{Liantao Ma}, \bibinfo{person}{Yinghao Zhu}, \bibinfo{person}{Yasha Wang}, \bibinfo{person}{Jiangtao Wang}, {and} \bibinfo{person}{Junfeng Zhao}.} \bibinfo{year}{2022}\natexlab{}.
\newblock \showarticletitle{M3care: Learning with missing modalities in multimodal healthcare data}. In \bibinfo{booktitle}{\emph{Proceedings of the 28th ACM SIGKDD conference on knowledge discovery and data mining}}. \bibinfo{pages}{2418--2428}.
\newblock


\bibitem[Zhang et~al\mbox{.}(2023)]%
        {zhang2023transformehr}
\bibfield{author}{\bibinfo{person}{Chao Zhang}, \bibinfo{person}{Yanbo Xu}, \bibinfo{person}{Shangqing Xu}, \bibinfo{person}{Manav Ramprassad}, {and} \bibinfo{person}{Alexey Tumanov}.} \bibinfo{year}{2023}\natexlab{}.
\newblock \showarticletitle{TransformEHR: Pretraining Transformer Models on Large-Scale EHR Data}.
\newblock \bibinfo{journal}{\emph{Nature Communications}} \bibinfo{volume}{14}, \bibinfo{number}{1} (\bibinfo{year}{2023}), \bibinfo{pages}{1234}.
\newblock


\bibitem[Zhao et~al\mbox{.}(2024)]%
        {zhao2024deep}
\bibfield{author}{\bibinfo{person}{Fei Zhao}, \bibinfo{person}{Chengcui Zhang}, {and} \bibinfo{person}{Baocheng Geng}.} \bibinfo{year}{2024}\natexlab{}.
\newblock \showarticletitle{Deep multimodal data fusion}.
\newblock \bibinfo{journal}{\emph{ACM computing surveys}} \bibinfo{volume}{56}, \bibinfo{number}{9} (\bibinfo{year}{2024}), \bibinfo{pages}{1--36}.
\newblock


\end{thebibliography}

% \appendix
% \label{appendix}
% \section{Appendix}
% \subsection{Dataset Details}
% \label{appendix: dataset}
% Structured Patient Information: This modality includes static, tabular data aggregated for each hospital admission. It consists of patient demographics (e.g., age, gender), admission-level information (e.g., insurance type), and diagnostic codes (ICD-9 and ICD-10) summarizing the patient's condition upon admission.

% Time-Series Data: We extracted dynamic, time-stamped clinical measurements recorded throughout a patient's stay. This includes vital signs (e.g., heart rate, respiratory rate, blood pressure) from the chartevents table and laboratory test results (e.g., blood counts, chemistry panels) from the labevents table. These sequences capture the physiological trajectory of the patient.

% Unstructured Text Notes: This modality consists of de-identified free-text clinical notes from the noteevents table. For our analysis, we specifically focused on discharge summaries, which provide a comprehensive narrative of a patient's hospital course, including history, diagnostic findings, and treatment plans.

\end{document}